\documentclass[conference]{IEEEtran}
\IEEEoverridecommandlockouts
\usepackage{cite}
\usepackage{amsmath,amssymb,amsfonts}
\usepackage{algorithmic}
\usepackage{graphicx}
\usepackage{textcomp}
\usepackage{xcolor}
\usepackage{subcaption}
\usepackage{float}
\usepackage{blindtext}
\def\BibTeX{{\rm B\kern-.05em{\sc i\kern-.025em b}\kern-.08em
    T\kern-.1667em\lower.7ex\hbox{E}\kern-.125emX}}
\begin{document}

\title{Deep Ice Layer Tracking and Thickness Estimation using Fully Convolutional Networks
}

\author{
\IEEEauthorblockN{Debvrat Varshney, Maryam Rahnemoonfar*, Masoud Yari}
\IEEEauthorblockA{\textit{Computer Vision and Remote Sensing Laboratory} \\
\textit{University of Maryland Baltimore County}\\
Baltimore, MD, USA \\
dvarshney, maryam*, yari@umbc.edu \\
{\footnotesize \textsuperscript{*}Corresponding Author}}

\and
\IEEEauthorblockN{John Paden}
\IEEEauthorblockA{\textit{Center for Remote Sensing of Ice Sheets (CReSIS)} \\
\textit{University of Kansas Lawrence}\\
Lawrence, KS, USA\\
paden@ku.edu}
}

\maketitle

\begin{abstract}
Global warming is rapidly reducing glaciers and ice sheets across the world. Real time assessment of this reduction is required so as to monitor its global climatic impact. In this paper, we introduce a novel way of estimating the thickness of each internal ice layer using Snow Radar images and Fully Convolutional Networks. The estimated thickness can be used to understand snow accumulation each year. To understand the depth and structure of each internal ice layer, we perform multi-class semantic segmentation on radar images, which hasn't been performed before. As the radar images lack good training labels, we carry out a pre-processing technique to get a clean set of labels. After detecting each ice layer uniquely, we calculate its thickness and compare it with the processed ground truth. This is the first time that each ice layer is detected separately and its thickness calculated through automated techniques. Through this procedure we were able to estimate the ice-layer thicknesses within a Mean Absolute Error of approximately 3.6 pixels. Such a Deep Learning based method can be used with ever-increasing datasets to make accurate assessments for cryospheric studies. 

\end{abstract}


\section{Introduction} \label{section-introduction}
\begin{figure}[h!]
    \centering
    \includegraphics[scale=0.45]{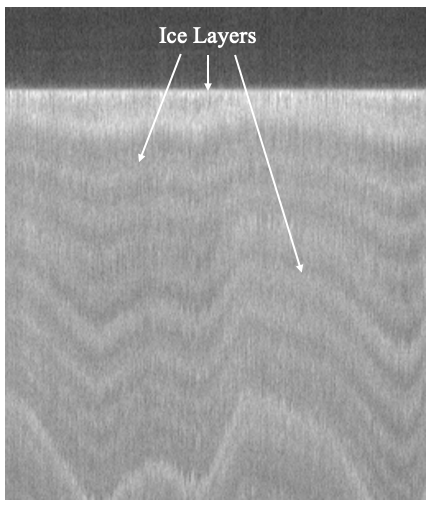}
    \caption{A sample snow radar image.}
    \label{fig:good_sample}
\end{figure}

Polar ice has been declining rapidly due to global warming. Studies suggest that sea level will increase by almost a meter at the end of this century \cite{IPCC2014}. To quantify and analyse this change, airborne Snow Radars \cite{snow-radar} are used which help in detecting internal ice-sheet layers. These instruments give two-dimensional grayscale images (Figure \ref{fig:good_sample} for example) where the horizontal axis corresponds to the flight direction of the instrument and the vertical axis corresponds to the depth in the sub-surface ice. The bright pixels correspond to signals reflected with a higher power, while the dark pixels correspond to signals reflected with a lower power \cite{cresis-2019}.
By analysing the depth of these ice layers, one can assess the snow accumulation rate\cite{koenig-accumulation-rate}. 


Snow radar imagery presents various challenges. Most of the radar images are noisy, with indistinguishable layers, especially the deeper layers. Moreover, the annotated layers are incomplete. As can be seen in Figure \ref{fig:good_sample} the lower layers of the snow radar image are thinner and have lesser contrast variation compared to the upper layers. This results in training labels being available (either through manual annotations or through traditional automated techniques) only for some parts of layers, and not for all parts of each layer. This is further explained in Section \ref{section-dataset}.

Detecting each layer separately is a challenge in itself. There have been several automated techniques in the past which detect ice layers from radar data \cite{bruzzone1, bruzzone2, level-set, charged-particles}. But these methods focused on a binary detection of an ice layer i.e. whether an ice layer is present at a pixel or not. The uncertainty in these binary outputs, along with the fact that an ice pixel could belong to just \textit{any} layer makes it very hard to calculate the depth of each layer uniquely even through post-processing. 


Recently, Convolutional Neural Networks (CNNs) have shown a lot of promise in understanding complex images and extracting features from them. They have especially been used for computer vision problems such as image classification, object detection, and semantic segmentation \cite{rahnemoonfar2018flooded, sheppard2017real, rahnemoonfar2017deep, rahnemoonfar2019discountnet, rahnemoonfar2019semantic}. CNNs contain convolving filters which can segregate the varying shapes and textures in an an image similar to how human vision works. CNNs have also been used to detect ice layers from radar images \cite{level-set,Rahnemoonfar_2019}. But, extracting each internal layer separately, and calculating its depth is still an issue. More recently, Fully Convolutional Networks (FCNs) \cite{fcn} have been introduced for semantically segmenting an entire image. By supplying enough diverse training labels to these networks, we can get pixel-wise classification of each image. HED \cite{hed}, a multi-scale FCN for edge detection, was used in \cite{yari-hed} to detect internal ice layers in Snow Radar data. However, the authors detected ice layers in a binary format, segmenting the image into ice-layer pixels, and non-ice-layer pixels. We aim to achieve a similar output but by detecting each layer (present at different depths) uniquely.

In this paper, we use some state-of-the-art FCNs to understand each internal ice layer uniquely, and thus semantically segment Snow Radar images. We do so by first discarding the incomplete layer-labels and populating the complete layer-labels within the inter-layer regions. This will help us prepare training data where every pixel across the depth of each layer has a label. By pixel-wise annotating each layer uniquely, we can feed its specific features to an FCN for it to learn. The pixel-wise distribution of the labels in the FCN output can then help us estimate the thickness of each layer.

The rest of the paper is distributed in the following sections: Section II describes past work in ice-layer detection using radargrams, and also covers some state-of-the-art FCNs for Computer Vision in recent years. Section III describes the Snow Radar dataset that we use, and the challenges faced while detecting ice-layers from it. Section IV gives the Methodology, and highlights how we process the available training labels before feeding them to FCNs, the FCN architectures in detail, and the post-processing we carry out in order to obtain layered outputs. Section V explains the hyperparameters that we use with every architecture, and the evaluation metrics we use to assess their outputs. Section VI then quantifies the results, and also highlights qualitative results. We conclude the paper in Section VII.

\section{Related Work} \label{section-background}

Although there have been various automated techniques for binary detection of ice layers from radar images \cite{bruzzone-carrer, bruzzone1, level-set, charged-particles}, there is no technique to the best of our knowledge which creates a multi-class output for radar images taken over ice-sheets, especially using neural networks. In this section, we briefly describe existing techniques and highlight some state-of-the-art FCNs for semantic segmentation.

\subsection{Ice Layer Tracking Techniques}
Several automatic techniques are available for tracking the ice surface and bottom \cite{bruzzone-carrer,bruzzone1,level-set,charged-particles}. While \cite{bruzzone-carrer} focused on developing a hidden markov model to process planetary radargrams,  \cite{bruzzone1} coupled Steger and Weiner filtering with denoising methods to detect linear features from radar data acquired over icy regions. Further, \cite{level-set} used a level set approach to evaluate airborne radar imagery whereas \cite{charged-particles} used anisotropic diffusion followed by a contour detection model to identify ice and bedrock layers. These methods, although giving accurate results, resulted in a binary ice-layer detection, i.e. they detected the presence or non-presence of ice for each pixel. Moreover, these methods focused on detecting only the surface and bottom layers of the radargrams. 

Tracking the internal ice sheet layers is much more difficult since the layers are compact and too close to each other. Although there were several works in this field such as \cite{macgregor2015,koenig-accumulation-rate,dePaulOnana,bruzzone-carrer} which used automated techniques to detect internal layers, none of these methods used deep learning; and hence were not scalable for larger datasets. Several recent efforts \cite{yari-hed, rahnemoonfar2020deep, yari2020multi-scale, rahnemoonfar2020radar, Oluwanisola2020snow} applied multi-scale deep learning techniques to track and identify internal ice layers. Although these are very efficient methods, they perform binary detection of layers, i.e. they detect the presence or absence of ice at a given pixel. As snow gets accumulated over the years, forming a separate ice layer for each year, there lies a potential to detect \textit{which} layer an ice-pixel belongs to. We aim to solve this problem of tracking the compact, closely spaced, internal ice layers and identifying each layer uniquely. We use deep learning for its recent successes and scalability to large datasets. Since we wanted a pixel-wise distribution of each layer, we used FCNs for semantic segmentation which is described in the following subsection.


\subsection{Semantic Segmentation}

FCNs have been used extensively for semantic segmentation of images. The immense applicability of these networks and semantic segmentation in particular has resulted in it becoming a fundamental topic in Computer Vision \cite{deeplabv3+}.

The concept of semantic segmentation was introduced in \cite{fcn}, where the the terminating fully connected layers from popular networks like VGG and AlexNet were replaced with fully convolutional layers to bring pixel-wise  classification. Since then, these networks have further been modified by various strategies such as global or average pooling \cite{net-in-net}, batch normalization, different activation \cite{relu-compare} and loss functions \cite{loss1}, multi-scale architectures \cite{hed,rcf,pspnet}etc. A fusion of various training strategies have led to their success. In this section, we briefly describe some of the very successful semantic segmentation networks in recent times, explaining their utility. In the following section, we give the details of these network architectures.

\paragraph{UNet}
This network \cite{unet} contains a contracting path and an expansive path, which are almost symmetric to each other, forming a U-shaped architecture. High resolution features from the contracting path are concatenated with the upsampled outputs in the expansive path in order to obtain high localization. Such a network strategy, combined with data augmentation, turned out to be useful in biomedical image segmentation, especially for detecting tissue deformations. The success of UNet in biomedical images \cite{unet} led to its application and improvement for other domains such as remote sensing and autonomous driving. 

\paragraph{PSPNet}
In \cite{pspnet}, the authors observe that in a traditional FCN, most of the errors were due to a lack of global contextual relationship between different receptive fields. Hence, they introduced a pyramid pooling module which empirically turned out to be an effective global contextual prior. They start with global average pooling as a baseline for global contextual prior, and then fuse it with different sub-region context hierarchically so as to contain information from different scales. They refer to this hierarchical structure as a \textit{pyramid pooling module}. Such a network strategy turns out to be very useful for snow radar dataset, as not only do we need to extract gradient changes in the noise locally, we also need to uniquely identify each layer from a global perspective. More details of the architecture are described in Section \ref{section-methodology}.

\paragraph{DeepLabv3+}
This architecture \cite{deeplabv3+} uses the concept of spatial pyramid pooling and applies several parallel atrous convolutions at different dilation rates to build an encoder architecture. The output of this is further upsampled (decoded) to extract features at the image scale. Such a combination of pyramid pooling and encoder-decoder architecture helps in learning multi-scale contextual information  while also detecting sharp object boundaries through spatial information.

We learnt that spatial pyramid pooling, such as that used in PSPNet, helps in learning a global contextual prior, but it also misses out on detailed object information due to the pooling operation. Atrous convolutions, help in this regard by dilating the receptive field in a controlled manner before it is pooled. The subsampled pooled output can then be decoded to obtain sharper spatial information. This network strategy can help us get highly detailed information about each layer change in the Snow Radar data. This architecture is further explained in Section \ref{section-methodology}. 

\section{Dataset}\label{section-dataset}
\subsection{Characteristics}
We use the Snow Radar data from 2012 year, provided by the Center for Remote Sensing of Ice Sheets (CReSIS) \cite{cresis-ku} and having a resolution of 4~cm per pixel in the vertical direction. This is publicly available and consists of 2361 training images and 260 test images. We have used the output of a semi-supervised technique \cite{koenig-accumulation-rate} as the ground truth. In our ground-truth data, each unique ice-layer is marked as a separate class. 

\subsection{Challenges}
\begin{figure}[h!]
\centering
\begin{subfigure}{0.5\columnwidth}
  \centering
  \includegraphics[width=0.9\linewidth]{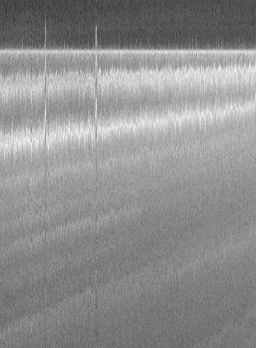}
  \caption{}
  \label{fig:sample_data_noisy_image}
\end{subfigure}%
\begin{subfigure}{0.5\columnwidth}
  \centering
  \includegraphics[width=0.9\linewidth]{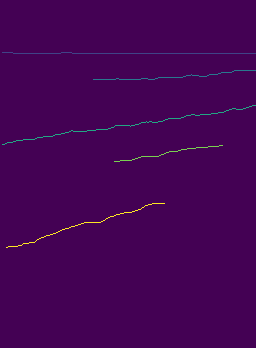}
  \caption{}
  \label{fig:sample_data_noisy_layer}
\end{subfigure}
\caption{Noisy radar image (a), having multiple indistinct layers, and its corresponding training labels (b). The lower layers of the radar image are not so easy to be distinguished by human eyes, whereas the available labels do not span across the corresponding layers completely.}
\label{fig:sample_data_noisy}
\end{figure}

The snow radar images are quite noisy, and it is very hard to distinguish where each layer begins. Moreover, there are hardly any contrasting features which can help us distinguish between layers. There are also certain anomalies, creating vertical perturbations in the horizontal ice layers. All these issues in the radar images can be seen in Figure \ref{fig:sample_data_noisy_image}.

These issues in the radar images propagate to the their labels, Figure \ref{fig:sample_data_noisy_layer}, which is the output of \cite{koenig-accumulation-rate} and which we use as ground truth. Most, if not all, the labels for the deeper layers are missing here. Moreover, the labels which are available, are incomplete as they do not cover the corresponding ice layer completely. These significant anomalies and issues in the original radar images, as well as the training labels, make it challenging for them to be directly trained with any FCN. As FCNs or CNNs are highly data dependent, any anomaly or issue in the original data or labels will directly propagate to the network output, leaving it to be of no practical use.

Thus, in order to get around these significant issues in the data and labels, we introduce some steps to process the training labels. These are described in Section \ref{label-processing} and help us in extracting only the complete labels for layers, by discarding the incomplete labels. We follow this procedure to crop out consecutive sets of complete training labels, and also crop out the corresponding regions from the original radar image. From the original CReSIS data of 2361 training images, our cropping procedure leaves us with 1157 images. 20\% of the these training images, i.e. 232, are explicitly used for neural network validation purposes. We use the entire 260 images for testing purposes, the annotations of which we were able to manually complete using the Darwin V7 platform \cite{V7Darwin}.


\section{Methodology} \label{section-methodology}
In this section, we discuss how we process the incomplete or missing training labels, the network architectures of the three FCNs, and the post-processing we carry out to obtain layered outputs. In the next section, we talk about the hyperparameter setup for the FCNs, and the evaluation metrics we use to assess their performance.

\subsection{Processing the Training Labels}\label{label-processing}

\begin{figure*}[h!]
\centering
\includegraphics[width=\textwidth]{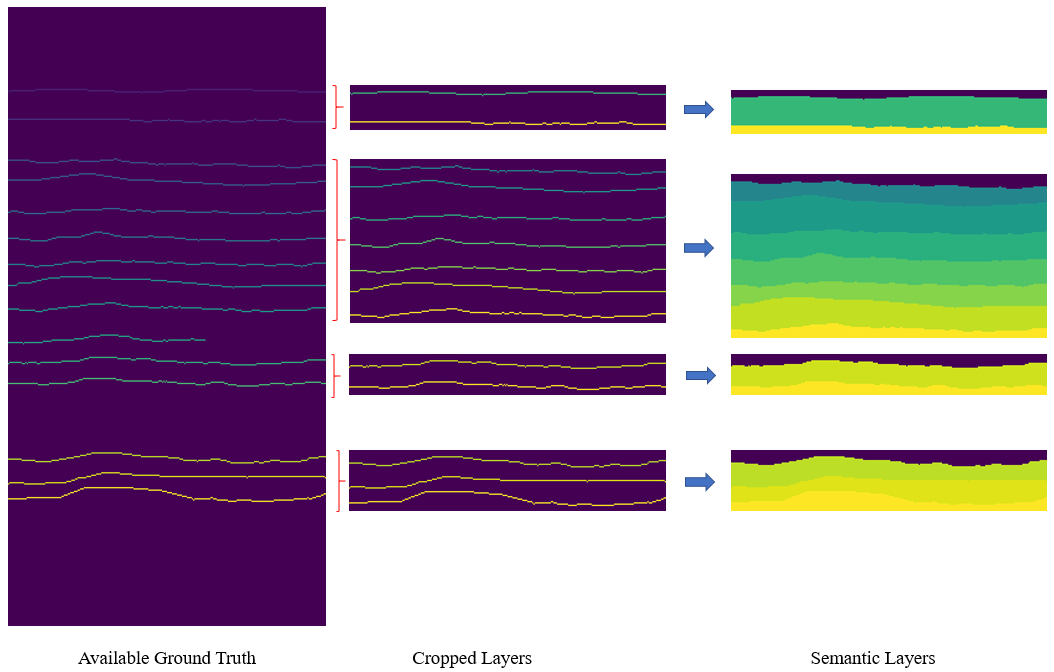}
\caption{Processing the Training Labels: First, consecutive sets of completely labeled layers are cropped out. In the second step, pixels in between two layers are filled up with the label of the upper layer. This pixel-wise annotation across the thickness of the layers will help FCN models to learn features of every layer uniquely. Each label color represents a unique layer. Note that the colors are generated in a spectrum, and are not necessarily consistent across the ground truth and the semantic output. The red curly braces represent the row indices for cropping.}
\label{fig:layer-cropping}
\end{figure*}

As some of the training labels available to us  were incomplete (such as the second, fourth and fifth layer in Figure \ref{fig:sample_data_noisy_layer}), we removed these completely, that is turn them into background pixels. Then, starting from the topmost layer, we searched for a consecutive set of at least two layers. For every consecutive set found, we calculate the row index of the peak of the top layer, and the row index of the valley of the bottom layer. We then added a margin of five to both these indices setting these as the \textit{y-} coordinate values of the bounding box for cropping this consecutive set of layers. The bounding box spans across all the columns of the training labels, i.e. its \textit{x-} coordinate values are the first and the last column index of the ground truth image respectively. These bounding box coordinates are then used to crop out the same region from the corresponding, original radar image.

Consider Figure \ref{fig:layer-cropping} as an example. The available ground truth here has the following layer-labels available: 2, 3, 5, 6, 7, 8, 9, 10, 11, 12, 13, 14, 18, 19, 20. There are no layers labelled as 1, 4, 15, 16, and 17. Also, layer labelled 12 is incomplete, and doesn't span across the width of the image. To process this ground truth, we first completely remove layer labelled 12, i.e. we turn it to background. We do this by converting all pixels having value 12, to have value 0 (the background class). So now we are left with layers labelled as: 2, 3, 5, 6, 7, 8, 9, 10, 11, 13, 14, 18, 19, 20. Out of these, the consecutive sets available are: \{2,3\}, \{5, 6, 7, 8, 9, 10, 11\}, \{13, 14\}, and \{18, 19, 20\}. The first layer in each set is its top layer, and the last layer in each set is its bottom layer. By calculating the row indices of the peaks of the top layers of each set, i.e. layers 2, 5, 13, and 18; and row indices of the valleys of the bottom layers of each set, i.e. layers 3, 11, 14, and 20; we are able to crop these sets out into separate images, after adding a margin of five pixels to the aforementioned row indices.

\begin{figure}[h!]
\centering
\includegraphics[width=0.5\textwidth]{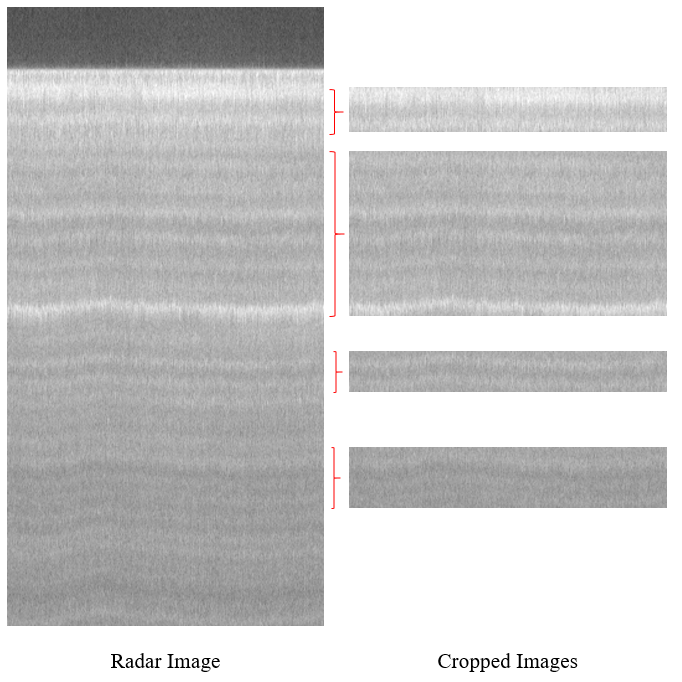}
\caption{This is the radar image for the training labels of Figure \ref{fig:layer-cropping}. This image is cropped at the same regions as its corresponding label-image in Figure \ref{fig:layer-cropping}. The red curly braces represent the row indices where it was cropped to generate `Cropped Images'.}
\label{fig:image-cropping}
\end{figure}

The bounding box coordinates are explained in Equation \ref{bbox}. For a peak row index $p$ and a valley row index $v$, of an image with width $w$, the bounding box used for cropping has coordinate values (the top-left coordinate, and the bottom-right coordinate) as computed by Equation \ref{bbox}. These same coordinates are used for cropping the corresponding radar image, as shown in Figure \ref{fig:image-cropping}.

\begin{equation}
    (\,x_1\,,\,y_1\,) \: , \: (\,x_2\,,\,y_2\,) = (\,0\,,\,p-5\,) \: , \: (\,w\,,\,v+5\,)
    \label{bbox}
\end{equation}

Further, in order to feed FCNs for semantic segmentation, we need training labels which are annotated for each pixel of the image. To accomplish this in the cropped subsets, we fill all the intermediate background pixels between two layers with the label of the \textit{upper} layer. This leaves us with labelled pixels across most of the image, except for the background pixels above the top-most layer of the image. This layer-filling process is also shown in Figure \ref{fig:layer-cropping}, where we generate the `Semantic Layers' from the `Cropped Layers'.

\subsection{Network Architectures}
We carried out semantic segmentation of Snow Radar images using three state-of-the-art FCNs: UNet \cite{unet}, PSPNet \cite{pspnet} and DeepLabv3+ \cite{deeplabv3+}. In this Section, we give some details about their architectures.

\subsubsection{UNet}
The architecture of UNet\cite{unet} consists of a contracting path and an expansive path. The contracting path (left side) has repeated applications of two 3$\times$3 unpadded convolutions and 2$\times$2 max pooling (of stride 2) operation for downsampling. Each convolution is followed by a rectified linear unit (ReLU) activation and at each downsampling step, the number of feature channels are doubled. The expansive path (right side) then focuses on upsampling the feature maps followed by 2$\times$2 convolutions which half the  number of channels, which are then concatenated with the corresponding cropped features maps from the contracting paths. These are then convolved by two 3$\times$3 filters each having a ReLU activation function. Finally, a 1$\times$1 convolution is used to reduce the feature vector to the desired number of classes. 
\subsubsection{PSPNet}
The pyramid pooling module fuses features from four different pyramid scales 
. The coarsest level generates a single bin output through a global pooling scheme, whereas other levels generate pooled representations for different sub-regions. These low dimension pooled outputs from different levels are then upsampled to get feature maps of the same size as the original feature map via bilinear interpolation. These different features are then concatenated to give the final prediction. The pyramid pooling module that we adopt has four bin size of 1$\times$1, 2$\times$2, 3$\times$3 and 6$\times$6 respectively. 

The baseline CNN that we use in the PSPNet architecture is ResNet-50 \cite{resnet}. However, contrary to the ResNet architecture, PSPNet incorporates an additional, auxiliary loss after the fourth stage (residual block) of ResNet to deeply supervise \cite{dsn} the network architecture. The entire network is trained by a weighted loss that balances between this auxiliary loss and the main, terminal loss.

\subsubsection{DeepLabv3+}
We build DeepLabv3+ \cite{deeplabv3+} using a ResNet-50\cite{resnet} as the baseline. We then apply multiple atrous convolutions with different dilation rates (6, 12 and 18) to extract the spatial information. This is then fused with the pooled output of the feature maps and later convolved with 1$\times$1 filters. These encoder features are then bilinearly upsampled by a factor of 4, to be later concatenated with the low-level features from the network backbone which have the same spatial resolution. Further, in the decoder path, 1$\times$1 convolutions are applied on these features to reduce the number of channels and make the training easier. Feature maps are then concatenated and convolved with a couple of 3$\times$3 filters to refine the features, which are then upsampled by a factor of 4 by using bilinear interpolation.

\subsection{Processing the Network Outputs}
The fully convolutional networks described above are expected to give us pixel-wise outputs, like those shown in `Semantic Layers' column of Figure \ref{fig:layer-cropping}. In order to convert them back to individual layers (such as `Cropped Layers' of Figure \ref{fig:layer-cropping}), we iterate over each row of every column of the output to convert all \textit{duplicate} labels to the background pixel. Thus, each column of the output will have only one unique pixel for every label. We do this for all the columns, thus re-constructing layered output similar to the `Cropped Layers' column of Figure \ref{fig:layer-cropping}.  

\section{Experimental Setup}\label{section-experiments}
This section explains the setup of our experiments and describes the metrics we used to assess the FCNs' performance to segment the images and calculate the ice-layer thickness. The maximum number of unique layers we had available were 27, and we trained all networks on 28 classes to predict layer pixels as well as background pixels.

In order to understand the usability of our networks especially for internal ice layer tracking, we calculate the metrics on the test images having more than 1, and more than 3 layers. As there were a lot of images with less than 3 layers, this test will give us an idea of how the networks perform for \textit{deeper} layers. Furthermore, as each layer corresponds to the snow accumulated in a particular year, we calculate the performance on the top 10 layers of the test set. This is to study the changes in the past decade which can help us predict any changes in the near future. 

\subsection{Hyperparameters}
\begin{figure}[h!]
\centering
\begin{subfigure}{.9\columnwidth}
  \centering
  \includegraphics[width=0.98\columnwidth]{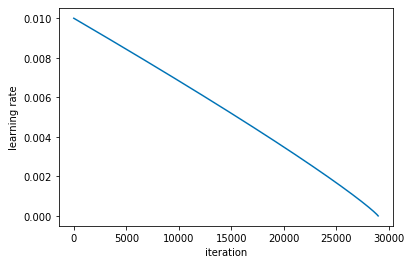}
  \caption{}
  \label{fig:lrs-poly}
\end{subfigure}
\begin{subfigure}{.9\columnwidth}
  \centering
  \includegraphics[width=0.98\columnwidth]{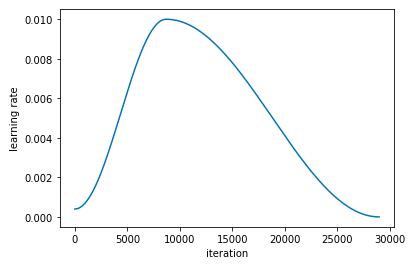}
  \caption{}
  \label{fig:lrs-onecycle}
\end{subfigure}%
\caption{(a) Poly Learning Rate Scheduler and (b) One Cycle Learning Rate Scheduler.} 
\label{fig:lrs}
\end{figure}

All the networks were trained with ResNet-50\cite{resnet} as the baseline network, and an initial learning rate of 0.01. We used a weight decay of $10^{-4}$ and a momentum of 0.9. We performed two types of experiments for each network, one where the learning rate scheduler was 'Poly' and the other where it was 'OneCycle'. In the Poly learning rate scheduler, the learning rate is linearly reduced from the initial value (0.01) to zero as the training progresses (Figure \ref{fig:lrs-poly}), while the momentum remains constant at 0.9. In the OnceCycle learning rate scheduler, the learning rate is annealed according to the one-cycle learning rate policy \cite{cos-annealing} (Figure \ref{fig:lrs-onecycle}). This means that, during the initial 30\% of the training, the learning rate increases from a tenth of its (0.001) value to its full value (0.01). For the remainder of the training, the learning rate decreases from its full value to a quarter of its value. The momentum is also changed in a similar but opposite way. It decreases from 0.9 to 0.8, and then later increases to 0.9 again during second half (remaining 70\%) of the training. All networks were trained with a cross entropy loss having a mini-batch size of 8 images for 200 epochs. 

\subsection{Evaluation Metrics}
We assess the performance of these networks with overall accuracy and mean IoU (intersection over union) per label. For $k$ labels, these metrics are given as

\begin{equation}
Accuracy = \frac{\sum\limits_{i=1}^k \frac{TP_i + TN_i}{TP_i + TN_i + FP_i + FN_i}}{k}
\label{eq:accuracy}
\end{equation}
\begin{equation}
mean\:IoU = \frac{\sum\limits_{i=1}^k \frac{Predicted\:Output_i\;\cap\; Ground\:Truth_i}{Predicted\:Output_i\;\cup\:Ground\:Truth_i}}{k}
\label{eq:mIoU}
\end{equation}

where TP, TN, FP, FN are True Positives, True Negatives, False Positives, and False Negatives, respectively. This is done on pixel-wise (semantic) outputs and ground truth in the format of `Semantic Layers' of Figure \ref{fig:layer-cropping}. 

We also calculate the \textit{mean} thickness of each layer in every predicted image, and compare it with the corresponding ground truth semantic layers. For calculating this mean thickness, we first count the total number of pixels for each unique class, and divide it by the number of columns (width) of the image. We then calculate the Mean Absolute Error (MAE) between the predicted output and the ground truth across all the layers of a given image. This is given by
\begin{equation}
MAE = \frac{\sum\limits_{i=1}^k\mid p_i-t_i \mid}{k}
\label{eq:mae}
\end{equation}
where $p_i$ is the predicted mean thickness and $t_i$ is the true mean thickness of the $i^{th}$ layer. 

\section{Results and Discussion}

\begin{figure*}
\centering

\begin{subfigure}{.2\textwidth}
  \centering
  \includegraphics[width=0.98\linewidth]{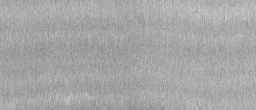}
  \caption*{}
\end{subfigure}%
\begin{subfigure}{.2\textwidth}
  \centering
  \includegraphics[width=0.98\linewidth]{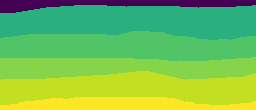}
  \caption*{}
\end{subfigure}%
\begin{subfigure}{.2\textwidth}
  \centering
  \includegraphics[width=0.98\linewidth]{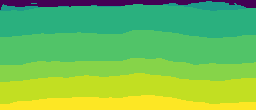}
  \caption*{}
\end{subfigure}%
\begin{subfigure}{.2\textwidth}
  \centering
  \includegraphics[width=0.98\linewidth]{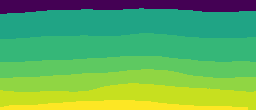}
  \caption*{}
\end{subfigure}%
\begin{subfigure}{.2\textwidth}
  \centering
  \includegraphics[width=0.98\linewidth]{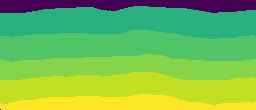}
  \caption*{}
\end{subfigure}

\begin{subfigure}{.2\textwidth}
  \centering
  \includegraphics[width=0.98\linewidth]{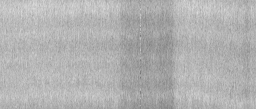}
  \caption*{}
\end{subfigure}%
\begin{subfigure}{.2\textwidth}
  \centering
  \includegraphics[width=0.98\linewidth]{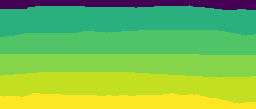}
  \caption*{}
\end{subfigure}%
\begin{subfigure}{.2\textwidth}
  \centering
  \includegraphics[width=0.98\linewidth]{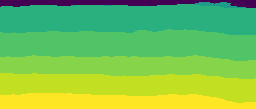}
  \caption*{}
\end{subfigure}%
\begin{subfigure}{.2\textwidth}
  \centering
  \includegraphics[width=0.98\linewidth]{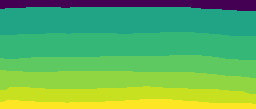}
  \caption*{}
\end{subfigure}%
\begin{subfigure}{.2\textwidth}
  \centering
  \includegraphics[width=0.98\linewidth]{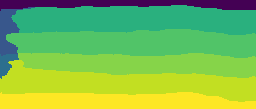}
  \caption*{}
\end{subfigure}

\begin{subfigure}{.2\textwidth}
  \centering
  \includegraphics[width=0.98\linewidth]{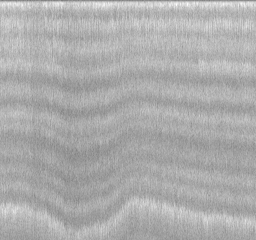}
  \caption*{}
\end{subfigure}%
\begin{subfigure}{.2\textwidth}
  \centering
  \includegraphics[width=0.98\linewidth]{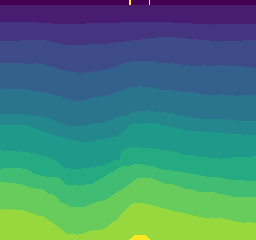}
  \caption*{}
\end{subfigure}%
\begin{subfigure}{.2\textwidth}
  \centering
  \includegraphics[width=0.98\linewidth]{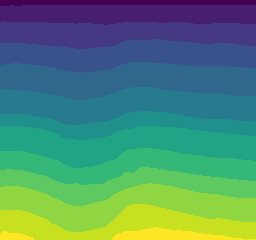}
  \caption*{}
\end{subfigure}%
\begin{subfigure}{.2\textwidth}
  \centering
  \includegraphics[width=0.98\linewidth]{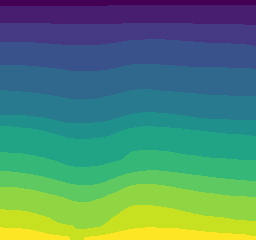}
  \caption*{}
\end{subfigure}%
\begin{subfigure}{.2\textwidth}
  \centering
  \includegraphics[width=0.98\linewidth]{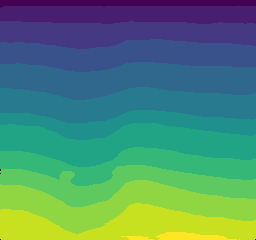}
  \caption*{}
\end{subfigure}

\begin{subfigure}{.2\textwidth}
  \centering
  \includegraphics[width=0.98\linewidth]{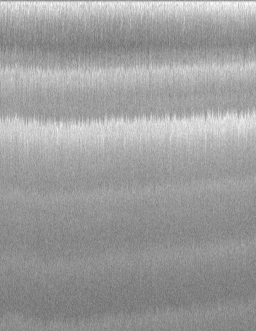}
  \caption*{}
\end{subfigure}%
\begin{subfigure}{.2\textwidth}
  \centering
  \includegraphics[width=0.98\linewidth]{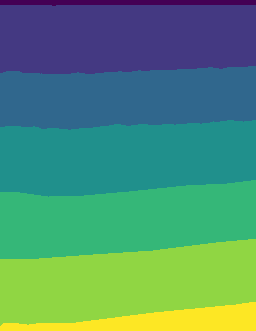}
  \caption*{}
\end{subfigure}%
\begin{subfigure}{.2\textwidth}
  \centering
  \includegraphics[width=0.98\linewidth]{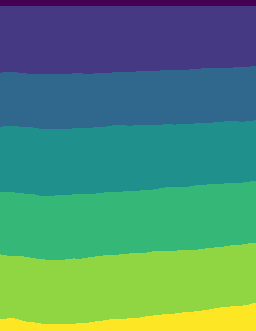}
  \caption*{}
\end{subfigure}%
\begin{subfigure}{.2\textwidth}
  \centering
  \includegraphics[width=0.98\linewidth]{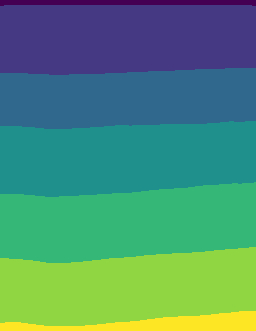}
  \caption*{}
\end{subfigure}%
\begin{subfigure}{.2\textwidth}
  \centering
  \includegraphics[width=0.98\linewidth]{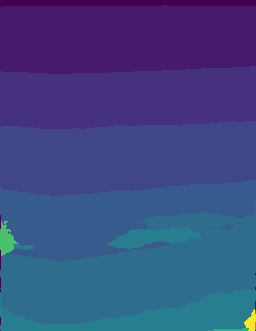}
  \caption*{}
\end{subfigure}

\begin{subfigure}{.2\textwidth}
  \centering
  \includegraphics[width=0.98\linewidth]{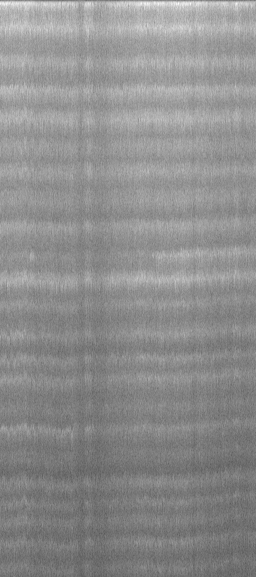}
  \caption*{Radar Image}
\end{subfigure}%
\begin{subfigure}{.2\textwidth}
  \centering
  \includegraphics[width=0.98\linewidth]{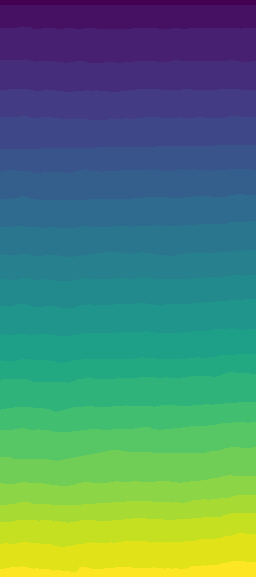}
  \caption*{(Semantic) Ground Truth}
\end{subfigure}%
\begin{subfigure}{.2\textwidth}
  \centering
  \includegraphics[width=0.98\linewidth]{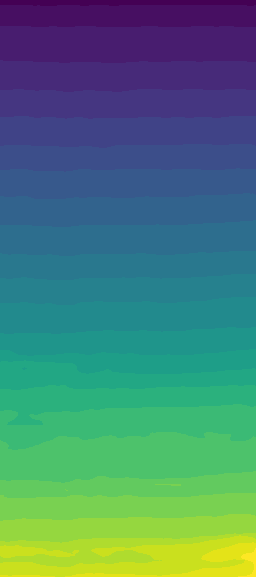}
  \caption*{DeepLabv3+}
\end{subfigure}%
\begin{subfigure}{.2\textwidth}
  \centering
  \includegraphics[width=0.98\linewidth]{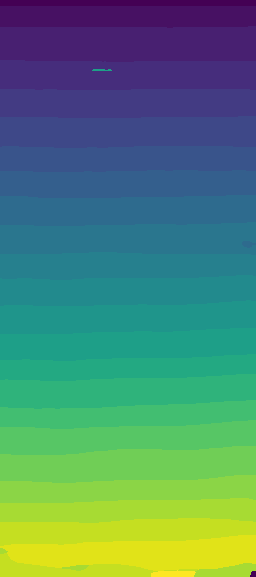}
  \caption*{PSPNet}
\end{subfigure}%
\begin{subfigure}{.2\textwidth}
  \centering
  \includegraphics[width=0.98\linewidth]{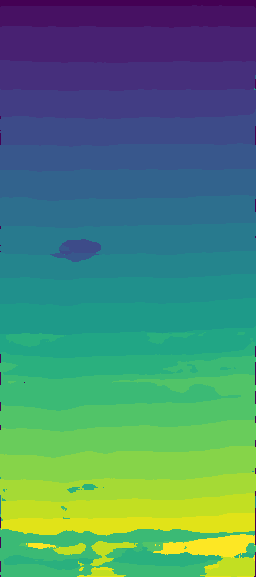}
  \caption*{UNet}
\end{subfigure}
\caption{Comparing the outputs of DeepLabv3+, PSPNet and UNet with respect to the test images and available ground truth.}
\label{fig:qualitative-semantic-comparison}
\end{figure*}

\begin{figure*}
\centering
\begin{subfigure}{.2\textwidth}
  \centering
  \includegraphics[width=0.98\linewidth]{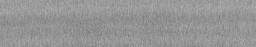}
  \caption*{}
\end{subfigure}%
\begin{subfigure}{.2\textwidth}
  \centering
  \includegraphics[width=0.98\linewidth]{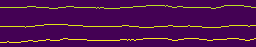}
  \caption*{}
\end{subfigure}%
\begin{subfigure}{.2\textwidth}
  \centering
  \includegraphics[width=0.98\linewidth]{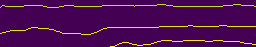}
  \caption*{}
\end{subfigure}%
\begin{subfigure}{.2\textwidth}
  \centering
  \includegraphics[width=0.98\linewidth]{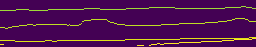}
  \caption*{}
\end{subfigure}%
\begin{subfigure}{.2\textwidth}
  \centering
  \includegraphics[width=0.98\linewidth]{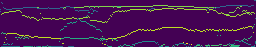}
  \caption*{}
\end{subfigure}

\begin{subfigure}{.2\textwidth}
  \centering
  \includegraphics[width=0.98\linewidth]{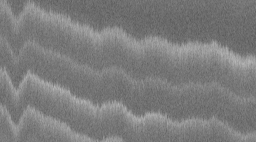}
  \caption*{}
\end{subfigure}%
\begin{subfigure}{.2\textwidth}
  \centering
  \includegraphics[width=0.98\linewidth]{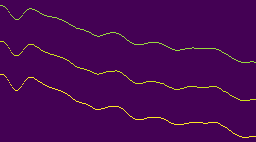}
  \caption*{}
\end{subfigure}%
\begin{subfigure}{.2\textwidth}
  \centering
  \includegraphics[width=0.98\linewidth]{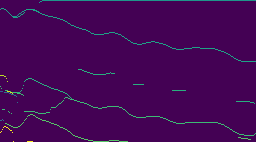}
  \caption*{}
\end{subfigure}%
\begin{subfigure}{.2\textwidth}
  \centering
  \includegraphics[width=0.98\linewidth]{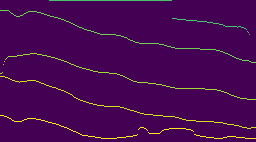}
  \caption*{}
\end{subfigure}%
\begin{subfigure}{.2\textwidth}
  \centering
  \includegraphics[width=0.98\linewidth]{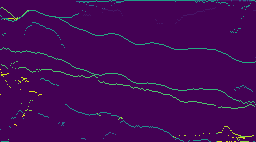}
  \caption*{}
\end{subfigure}

\begin{subfigure}{.2\textwidth}
  \centering
  \includegraphics[width=0.98\linewidth]{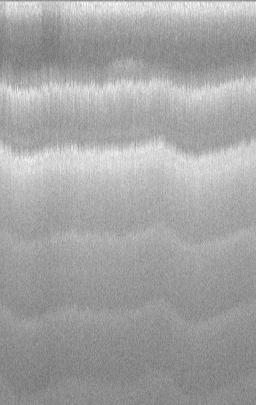}
  \caption*{}
\end{subfigure}%
\begin{subfigure}{.2\textwidth}
  \centering
  \includegraphics[width=0.98\linewidth]{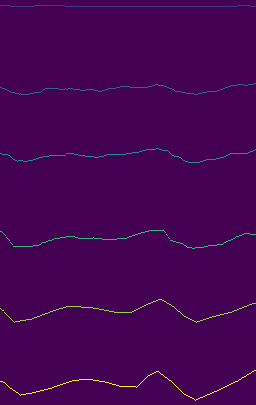}
  \caption*{}
\end{subfigure}%
\begin{subfigure}{.2\textwidth}
  \centering
  \includegraphics[width=0.98\linewidth]{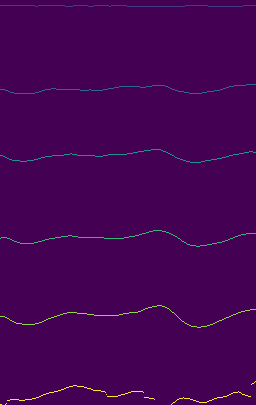}
  \caption*{}
\end{subfigure}%
\begin{subfigure}{.2\textwidth}
  \centering
  \includegraphics[width=0.98\linewidth]{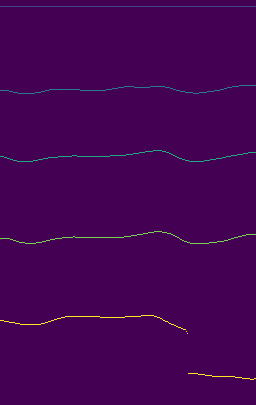}
  \caption*{}
\end{subfigure}%
\begin{subfigure}{.2\textwidth}
  \centering
  \includegraphics[width=0.98\linewidth]{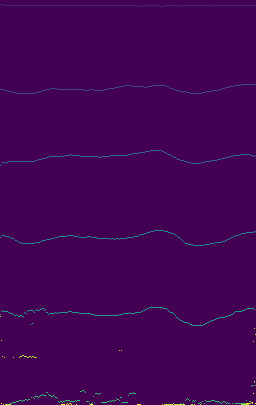}
  \caption*{}
\end{subfigure}

\begin{subfigure}{.2\textwidth}
  \centering
  \includegraphics[width=0.98\linewidth]{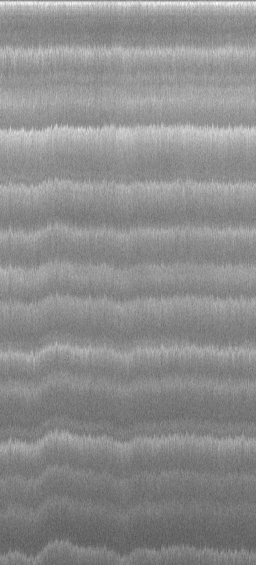}
  \caption*{Radar Image}
\end{subfigure}%
\begin{subfigure}{.2\textwidth}
  \centering
  \includegraphics[width=0.98\linewidth]{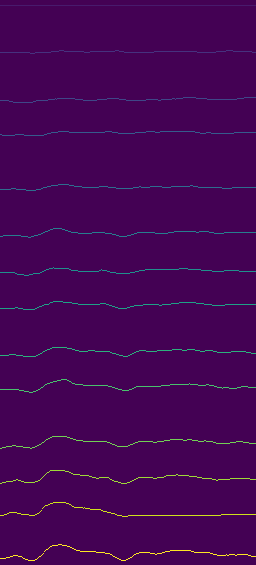}
  \caption*{Ground Truth}
\end{subfigure}%
\begin{subfigure}{.2\textwidth}
  \centering
  \includegraphics[width=0.98\linewidth]{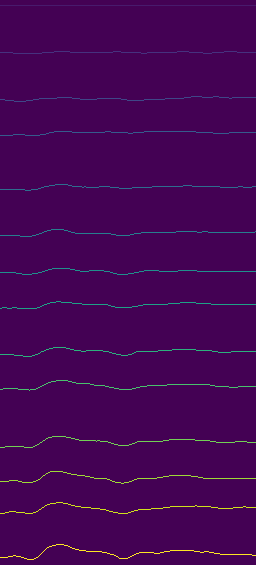}
  \caption*{DeepLabv3+}
\end{subfigure}%
\begin{subfigure}{.2\textwidth}
  \centering
  \includegraphics[width=0.98\linewidth]{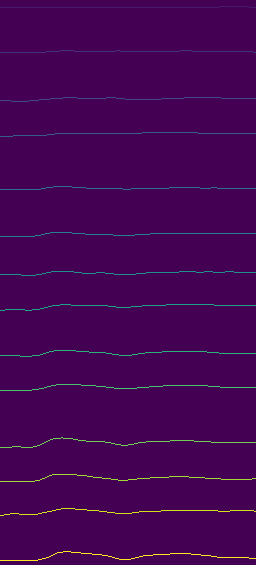}
  \caption*{PSPNet}
\end{subfigure}%
\begin{subfigure}{.2\textwidth}
  \centering
  \includegraphics[width=0.98\linewidth]{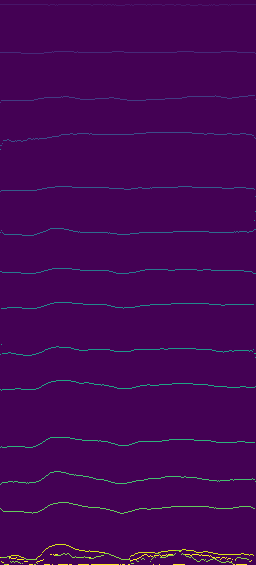}
  \caption*{UNet}
\end{subfigure}
\caption{Comparing the outputs of DeepLabv3+, PSPNet and UNet with respect to the test images and available ground truth.}
\label{fig:qualitative-comparison}
\end{figure*}



\begin{table}[h!]
\centering
\begin{tabular}{l l l l} 
 Network-LRS & Train & Val & Test \\ 
 \hline
 UNet-Poly & 0.755 & 0.681 & 0.714 \\ 
 UNet-OneCycle & 0.856 & 0.754 & 0.792 \\
 PSPNet-Poly & 0.948 & 0.875 & \textbf{0.899} \\ 
 PSPNet-OneCycle & 0.938 & 0.844 & 0.867 \\ 
 DeepLabv3+-Poly & 0.957 & \textbf{0.907} & 0.887 \\  
 DeepLabv3+-OneCycle & 0.935 & 0.876 & 0.886 \\ 
\end{tabular}
\caption{Accuracy of various networks on the Training, Validation and Test set computed over all the 27 layers. LRS denotes the Learning Rate Strategy - i.e. Poly or OneCycle. The highest values obtained over the Validation and Test sets are highlighted in bold.}
\label{table:accuracy}
\end{table}

\begin{table}[h!]
\centering
\begin{tabular}{l l l l} 
 Network-LRS & Train & Val & Test \\ 
 \hline
 UNet-Poly & 0.387 & 0.288 & 0.343 \\ 
 UNet-OneCycle & 0.549 & 0.378 &  0.438 \\
 PSPNet-Poly & 0.737 & 0.576 & 0.65 \\ 
 PSPNet-OneCycle & 0.728 & 0.538 & 0.589 \\ 
  DeepLabv3+-Poly & 0.734 & \textbf{0.609} & 0.59 \\  
 DeepLabv3+-OneCycle & 0.676 & 0.552 & \textbf{0.595} \\ 
\end{tabular}
\caption{Mean Intersection over Union (IoU) of various networks on the Training, Validation and Test sets computed over all the 27 layers. LRS denotes the Learning Rate Strategy - i.e. Poly or OneCycle. The highest values obtained over the Validation and Test sets are highlighted in bold.}
\label{table:mIoU}
\end{table}

We calculated the accuracy and mean IoU per (layer) class based on Equations \ref{eq:accuracy} and \ref{eq:mIoU} respectively. We tabulate these for all six experiments (three networks, and two learning rate schedulers per network) in Tables \ref{table:accuracy} and \ref{table:mIoU}. In these tables we further highlight the highest accuracy and mean IoU obtained over Validation and Test sets obtained over all the 27 layers. We also calculated the performance metrics on the test images having more than 1 and more than 3 layers, with the metrics calculated over the top 10 layers; which is more relevant for climate studies. These are shown in Tables \ref{tab:top10_test_accuracy} and \ref{tab:top10_test_mIoU}.

\begin{table}[h!]
    \centering
    \begin{tabular}{l l l}
         Network-LRS & $>$ 1 layers & $>$ 3 layers \\ 
         \hline
         UNet-Poly & 0.633 & 0.778 \\ 
         UNet-OneCycle & 0.626 & 0.82 \\
         PSPNet-Poly & 0.797 & 0.947 \\ 
         PSPNet-OneCycle & 0.681 & 0.888 \\ 
          DeepLabv3+-Poly & 0.733 & 0.915 \\  
         DeepLabv3+-OneCycle & \textbf{0.831} & \textbf{0.943} \\
    \end{tabular}
    \caption{The accuracy calculated over the top 10 layers of the test set. The results here are for images having more than 1, and more than 3 internal layers, which is more useful for our analysis.}
    \label{tab:top10_test_accuracy}
\end{table}

\begin{table}[h!]
    \centering
    \begin{tabular}{l l l}
         Network-LRS & $>$ 1 layers & $>$ 3 layers \\ 
         \hline
         UNet-Poly & 0.266 & 0.302 \\ 
         UNet-OneCycle & 0.292 & 0.351 \\
         PSPNet-Poly & 0.387 & 0.419 \\ 
         PSPNet-OneCycle & 0.366 & 0.413 \\ 
          DeepLabv3+-Poly & 0.392 & 0.424 \\  
         DeepLabv3+-OneCycle & \textbf{0.425} & \textbf{0.435} \\
    \end{tabular}
    \caption{The mean IoU calculated over the top 10 layers of the test set. The results here are for images having more than 1, and more than 3 internal layers, which is more useful for our analysis.}
    \label{tab:top10_test_mIoU}
\end{table}

From tables \ref{table:accuracy} and \ref{table:mIoU} we see that, the Poly learning rate gives higher performance with the multi-scale networks of PSPNet and DeepLabv3+. For UNet, the OneCycle learning rate works better. The gradual increase in the learning rate does not help with the multiple pooling strategies that PSPNet and DeepLabv3+ incorporate. For DeepLabv3+, both the learning rate schedulers give similar accuracy and mean IoU over the test set. Overall, DeepLabv3+ gave a higher mean IoU over the 27-layered Validation and Test sets, while also giving the highest accuracy over the Validation set. UNet performed the worst, both qualitatively (Figures \ref{fig:qualitative-semantic-comparison} and \ref{fig:qualitative-comparison}) and quantitatively (Tables \ref{table:accuracy} and \ref{table:mIoU}). For most of the images in Figure \ref{fig:qualitative-semantic-comparison}, UNet creates botchy patches, not being able to predict a layer completely across its width. This further leads to a lot of broken lines when we convert the semantic outputs to layered outputs (Figure \ref{fig:qualitative-comparison}).

We believe that the poor performance of UNet is due to its primitive architecture, as compared to PSPNet and DeepLabv3+. DeepLabv3+ captures not only a global contextual prior, but it is also able to retain intricate spatial information. Due to these reasons, it is able to decipher the highly ambiguous ice-layers while detecting them from a broader perspective. 


\begin{table}[h!]
\centering
\begin{tabular}{l l l l} 
 Network-LRS & Train & Val & Test \\ 
 \hline
 UNet-Poly & 7.95 & 10.14 & 8.75 \\ 
 UNet-OneCycle & 5.22 & 7.66 & 6.17 \\
 PSPNet-Poly & 2.80 & 4.79 & 3.63 \\ 
 PSPNet-OneCycle & 4.03 & 7.34 & 5.62 \\ 
 DeepLabv3+-Poly & 2.36 & \textbf{3.66} & 3.75 \\  
 DeepLabv3+-OneCycle & 3.08 & 4.53 & \textbf{3.59} \\  
\end{tabular}
\caption{The Mean Absolute Error (MAE) in pixels of all the layers calculated over the Training, Validation and Test sets. Values highlighted in bold are the least thickness values obtained over the Validation and Test sets.}
\label{table:thickness-mae}
\end{table}

We also calculated the thickness of each layer in the networks' predicted output and compared it with the semantic ground truth layers (such as those present in Figure \ref{fig:layer-cropping}). We report the Mean Absolute Error (MAE, Equation \ref{eq:mae}) of all the layers across all the images of each dataset (i.e. Training, Validation and Test sets) in Table \ref{table:thickness-mae}. From this table, we see that the semantic segmentation networks have predicted well, resulting in an average MAE across all datasets and all networks to be 5.28 pixels. Further, DeepLabv3+ gave the best outputs with its MAE falling between 3 to 4 pixels for the Validation and Test sets; closely followed by PSPNet which gave an MAE between 3.5 and 7 pixels. UNet's outputs gave the worst thickness estimates, with an MAE of upto 10 pixels. This is majorly due to the botchy semantic output that it generates over the deeper layers. 

Although training till 200 epochs could not improve UNet's output, tuning other hyperparameters apart from learning rate should definitely improve its results. Further, as UNet is a 'lighter' model in terms of number of weights, complexity, and mathematical operations involved, experimenting with it would be useful. 


\section{Conclusion}

Global warming is rapidly melting glaciers and ice sheets across the world. This calls for automated accurate methods which can process the large amount of data that is available from Earth observation. In this paper, we use Snow Radar data to track internal ice sheets and estimate the thickness of each layer. More specifically, we mitigate some of the challenges of the data set and its ground truth by a set of pre-processing techniques. We also use state-of-the-art fully convolutional networks to understand the pixel-wise distribution and extent of each ice-layer. By using this methodology, we are able to estimate the thickness of these layers within a Mean Absolute Error of 3 to 4 pixels. 

Even a slight change in the polar ice-sheets can be devastating for the world. Our work can thus be expanded by incorporating datasets from multiple years, and creating data-driven, real-time monitoring solutions which can go beyond label ambiguities.



\section*{Acknowledgment}
This work is supported by NSF BIGDATA awards (IIS-1838230, IIS-1838024), IBM, and Amazon.
\bibliographystyle{IEEEtran}
\bibliography{main}
\end{document}